\pdfoutput=1
\documentclass[runningheads]{llncs}
\usepackage[T1]{fontenc}
\usepackage{lmodern}
\usepackage{graphicx}
\usepackage{booktabs}
\usepackage{amsmath,amssymb}
\usepackage{multirow}
\usepackage{array}
\usepackage[hidelinks]{hyperref}
\usepackage{microtype}
\begin{document}
\title{Agent Benchmarks Stop Ranking Before They Run Out of Tasks}
\titlerunning{Agent Benchmarks Stop Ranking Before They Run Out of Tasks}
\author{Zihan Dong\inst{1} \and Qishi Zhan\inst{2} \and Minxuan Hu\inst{3}}
\authorrunning{Z. Dong et al.}
\institute{Lenovo Group\\ \email{puma122707@gmail.com} \and
Marquette University \and
Cornell University}
\maketitle

\begin{abstract}
Data-science and ML-engineering agents are evaluated on benchmarks such as MLE-bench, ALE-Bench, SWE-fficiency and AlgoTune, whose tasks yield a quantity (a Kaggle metric, contest score or speedup). When one stops separating the best models, the usual remedy is a harder version. Re-scoring released runs of eight benchmarks, including SWE-bench Verified and Terminal-Bench for breadth, we find much of this saturation lies in the measurement, not the tasks. Most agent benchmarks reduce a run to one pass/fail bit, which fails in two ways. It is too coarse: neighbouring top-ten agents on SWE-bench Verified differ by less than its tests' errors, and correcting them reorders neighbours no more than deleting a random sixth of each agent's passes. It is capped at the reference: ALE-Bench's top ten, separated by contest score, tie when graded against the median human, although nearly half the runs beat that human. On fixed tasks, a higher bar only moves the problem; graded credit fixes it. We derive a clipped log-ratio score with dated anchors and cost as a separate axis. In a Terminal-Bench re-run, pass/fail-tied models differ over tenfold on the quantity the tests discard.
\end{abstract}

\section{Introduction}\label{sec:intro}

Data-science and ML-engineering agents are benchmarked on Kaggle competitions (MLE-bench~\cite{mlebench2024}), optimization contests (ALE-Bench~\cite{alebench2025}) and code speedups (SWE-fficiency~\cite{swefficiency2025}, AlgoTune~\cite{algotune2025}). Consider ALE-Bench, 40 contests scored by contest performance: its top ten separate into three tiers. Scored pass/fail against the median human, as most agent benchmarks score runs, five models pass 39 of the 40 problems and the top ten become one tier, although nearly half the runs beat that human. On the same tasks, a higher bar does not fix this: one that separates the newest models scores 89\% of the earlier ones zero. A single pass/fail bar per task sees one band of ability; graded credit widens it.

\begin{figure}[tb]
\centering
\includegraphics[width=\textwidth]{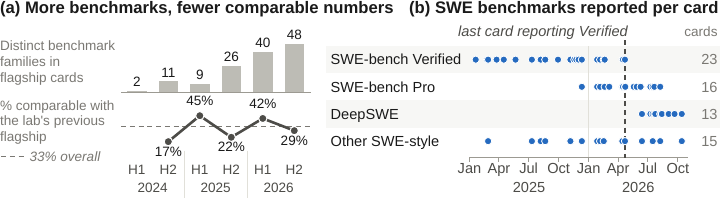}
\caption{Benchmark turnover in 57 flagship model cards (2024--2026). \textbf{(a)} Agentic benchmark families reported per half-year, and the share of a card's agentic results comparable with the lab's previous flagship. \textbf{(b)} Which SWE-style benchmark each card reports.}
\label{fig:census}
\end{figure}

The symptom is general: the last flagship model card to report SWE-bench Verified appeared in April 2026 (Fig.~\ref{fig:census}b), when its top ten solved 76--80\% of the tasks: it ran out of \emph{ranking}, not tasks. Across 57 flagship cards from seven labs, only a third of a card's agentic results are comparable with the same lab's previous flagship (same benchmark, version, subset and harness; Fig.~\ref{fig:census}a).

Turnover has many causes, contamination among them; we use it as motivation only. The usual reading of saturation is that the tasks became too easy, and the usual remedy is a harder version. We find that much of it lies in the \emph{measurement}. Almost every agent benchmark grades a run against an \emph{answer key} written for each task (a reference solution, an expected end state, or tests derived from the author's solution), which returns one bit. The key is also the costliest and least verifiable part of a benchmark: a checker that reliably separates good answers from bad ones is case-specific, slow to write and still often wrong~\cite{swebenchplus2024,abc2025,misscore2026}, and OSWorld reports twice as much effort per task on the evaluator as on the task~\cite{osworld2024}. Where no key can be written, the fallback is an LLM judge, which a single token can fool~\cite{onetoken2025}.

The answer key thus fails in two separate ways. \textbf{Failure A, the bit is too coarse (§\ref{sec:c1}):} on three leaderboards, neighbouring agents in the top ten differ by less than a few hundred bits can detect, and on SWE-bench Verified by less than the tests' errors. \textbf{Failure B, the bit is capped (§\ref{sec:c2}--\ref{sec:score}):} a run that beats the reference, or passes at a fifth of the cost, scores the same as one that just passes.

Both failures come from one design choice: the key encodes an \emph{answer} rather than a \emph{preference}. An answer must be written for every task and stops informing once agents reach it. A preference (``faster is better'') can be scored once per domain, with room above any reference. Reinforcement learning made this switch when agents passed humans~\cite{mnih2015dqn}, as did optimization benchmarking~\cite{berthold2013primal,hansen2016coco}.

\paragraph{Contributions.}
(1)~On released runs of eight benchmarks and a Terminal-Bench re-run, we measure both failures, locate the lost information above the reference, and report where it does not help (§\ref{sec:c1}--\ref{sec:c3}).
(2)~We show why, on a fixed task set, a higher pass/fail bar cannot fix a saturated key while graded credit can, and derive the form of the graded score (§\ref{sec:score}).
(3)~We derive an evaluation methodology with a hardened, auditable gate, an anchored objective, cost as a separate reported axis, and frozen anchors that keep versions reproducible and comparable, instantiated in a small re-run; LLM judges help write evaluators but do not score (§\ref{sec:arch}). We release code and derived statistics.

\section{Related Work}\label{sec:related}

\paragraph{Saturation.} Saturation of 60 LM benchmarks has been tested against answer format, access, size and age~\cite{akhtar2026plateau}, not against how the score is built. Two 2026 analyses argue that hard answer keys become unaffordable or impossible once models exceed their authors~\cite{benchmarkceiling2026,edgecomprehension2026}. Evaluation results need error bars~\cite{miller2024errorbars,madaan2024variance}, and top SWE-bench submissions are statistically tied~\cite{converged2026}; we find the same for neighbouring agents on two more leaderboards and relate it to verifier error. Item response theory selects tasks of informative difficulty~\cite{polo2024tinybenchmarks}, and test equating links versions~\cite{kolen2014equating}; we test calibrated per-task bars and link versions through anchors. Surveys of agent evaluation report fragile evaluators, contamination and fast-moving versions~\cite{agencyeval2026}, and rankings are fragile to the wording of the task~\cite{wise2026}.

\paragraph{Verifiers and judges.} Weak tests let wrong patches pass on SWE-bench~\cite{swebenchplus2024,utboost2025,sweabs2026}, false fails are common~\cite{misscore2026}, and seven of ten audited agent benchmarks have flawed outcome checks~\cite{abc2025}. This work makes the bit valid and, in our design, hardens the \emph{gate}; we ask what the score should measure once it is.

\paragraph{Beyond pass/fail.} Some leaderboards report the dollar cost of each agent beside its accuracy~\cite{kapoor2024agents,hal2025,costofpass2025}; time horizons express success in human time~\cite{metr2025horizon}. Optimization benchmarks score an objective in single domains~\cite{alebench2025,algotune2025,swefficiency2025,mlebench2024}, and NPPC pairs one checker with generated instances~\cite{nppc2025}. These rules are not new; we show when they matter and why a new version cannot replace them.

\section{Failure A: The Bit Is Too Coarse to Rank the Frontier}\label{sec:c1}

A leaderboard ranks agents only if the gaps between neighbours exceed two errors: \emph{statistical} error from a finite task set, and \emph{verifier} error from a key that accepts wrong solutions or rejects right ones. For \textbf{SWE-bench Verified}~\cite{swebench2024} we use per-instance verdicts for the official top 30 (1~Sep 2026), whose passing patches SWE-ABS~\cite{sweabs2026} re-checked with strengthened tests. For \textbf{Terminal-Bench~2.0}~\cite{terminalbench2026} we use per-trial results of 61 submissions, after removing three flagged for cheating. For \textbf{OSWorld-Verified}~\cite{osworld2024} the top ten, seven with per-task trajectories.

We group agents into \emph{tiers}, sets of agents that the benchmark cannot tell apart, by walking down the ranking: an agent starts a new tier if it is significantly worse than the current tier's leader. As that leader is the best observed agent, we test all pairs at once, by a step-down max-$T$ test on paired $t$-statistics with a null from permuting agent labels within tasks~\cite{westfall1993resampling,romano2005exact}. This bounds the chance of any false split at 5\% (3--6\% for ten equal agents in simulation, against 11--24\% for Holm over the walk's own comparisons).

\begin{figure}[tb]
\centering
\includegraphics[width=\textwidth]{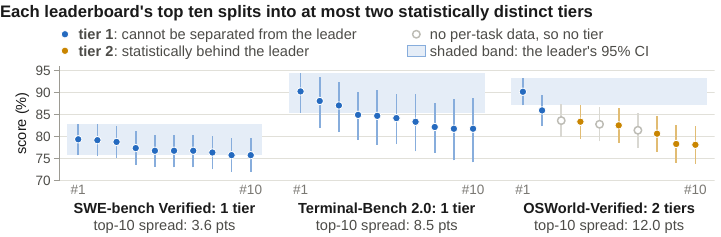}
\caption{The top ten agents on three leaderboards, with 95\% confidence intervals. Colours mark tiers; tier~1 cannot be separated from the leader. OSWorld-Verified tiers use the seven agents with per-task data.}
\label{fig:resolution}
\end{figure}

\paragraph{Neighbours at the top are statistically tied.} In September 2024, the SWE-bench Verified top ten spanned 19.2 points and three tiers; since March 2025 they have formed one or two (today 3.6 points, one tier; two in 59\% of task resamples). Today Terminal-Bench~2.0's top ten form one tier and OSWorld-Verified's two (Fig.~\ref{fig:resolution}). On SWE-bench Verified and Terminal-Bench~2.0, none of the 45 pairs within the top ten differs significantly (Holm over all 45), and on all three leaderboards no neighbouring pair does; on OSWorld-Verified a leading pair stands apart from a tied pack. The test is not weak: it detects about 6.3 points with 80\% power and splits ten real entries spaced 5.7 points apart into ten tiers, but the largest gap between SWE-bench neighbours is 1.4 points. Nor is it a ceiling: together the top ten solve 91\% of the tasks.

\paragraph{The tests' errors are larger than the gaps.} On SWE-bench Verified, strengthened tests reject 17.7\% of the median agent's passing patches (range 10.8--23.2\%). The rate differs between agents beyond sampling noise ($\chi^2_{29}=77.3$, permutation $p<10^{-4}$), with a spread of about 2 points, while neighbours are typically 0.2--0.4 points apart. Correcting the verdicts shrinks the top tier of the top 30 from 23 agents to a leading group of five, so a better key changes who is in it. The 18 agents that leave it stay nearly tied among themselves (1 of 153 pairs separates), and no neighbouring pair separates. The correction reverses 8 of the 18 neighbouring orders with a non-zero gap (Kendall $\tau=0.51$ between the two leaderboards), although in 11 of these pairs the two agents' inflation differs significantly and by more than the gap. Removing the same share of passes \emph{at random} reverses about as many (8; 95\% band 4--11). So the order of neighbours is noise, which any perturbation of this size reorders. False fails, about one in six failed runs in an OSWorld audit~\cite{misscore2026}, add to this error; none are released per agent, and simulated, they separate no neighbouring pair at a common rate (5--25\%) and at most 9\% of runs at agent-specific rates.

This failure has nothing to do with the cap. Its remedies are a stronger gate, more tasks or trials, more information per task, and tiers instead of ranks.

\section{Failure B: The Bit Is Capped at the Reference}\label{sec:c2}

If a key stops resolving the frontier because it is capped, the same runs scored with room above the cap should resolve it again, where the cap binds. We use every benchmark we found that releases per-run, per-task continuous results; 4\% to 71\% of their runs beat the reference (Table~\ref{tab:c2}).

Two of these benchmarks officially grade pass/fail per task: MLE-bench~\cite{mlebench2024} by medals, and METR~\cite{metr2025horizon} by binarized success. The other three officially score a continuous quantity; they let us ask what they would report had they been built as answer keys, as most agent benchmarks are. We score the same released runs three ways. The \emph{answer key} is the official binary verdict where one exists, and otherwise whether the run reaches the reference: the median human contestant on ALE-Bench~\cite{alebench2025}, the expert patch's speedup on SWE-fficiency~\cite{swefficiency2025}, and a 1.1-fold speedup over the reference solver on AlgoTune~\cite{algotune2025}. The \emph{capped} score gives partial credit up to the reference and none beyond it. The \emph{objective} is the score of §\ref{sec:score}: $0$ for runs that fail the benchmark's own validity check or do no better than a baseline, $1$ at the same reference, and above $1$ for runs that beat it, up to a clip at $3$. The baseline is the bottom-quartile human on ALE-Bench, no speedup on SWE-fficiency, a zero speedup on AlgoTune, and the median submission on MLE-bench; on AlgoTune, whose baseline is zero, $\log_2(1+\text{speedup})$ replaces the log-ratio.

We also report a paired variant of the saturation index of~\cite{akhtar2026plateau}, $S=\exp(-R^2)$, where $R$ is the mean gap between the first and fifth model over the standard deviation of their per-task differences, times $n^{1/4}$ for $n$ tasks: near $1$, the top five are within noise; near $0$, clearly separated.

\begin{table}[tb]
\caption{Same runs, scored by each benchmark's official metric and three ways of ours. ``Beat ref.'': share of runs above the objective's reference (median human, expert speedup, human baseliner, reference solver, gold medal). Cells: tiers among the top ten (more is a finer ranking), with $S$ in brackets (lower is less saturated). SWE-fficiency: seven models; 430 of 498 tasks, those with expert speedup $\ge 1.2$ (official score: all 498). $^\dagger$Officially pass/fail. $^*$Released capped at the reference.}\label{tab:c2}
\centering\small\setlength{\tabcolsep}{3pt}
\begin{tabular}{@{}lrrrcccc@{}}
\toprule
Benchmark & Models & Tasks & Beat ref. & Official & Answer key & Capped & Objective \\
\midrule
ALE-Bench     & 145 & 40  & 47.8\% & 3 (0.07) & 1 (1.00) & 1 (0.93) & 3 (0.03) \\
SWE-fficiency & 7   & 430 & 36.3\% & 1 (0.95) & 4 (0.10) & 5 (0.01) & 4 (0.02) \\
AlgoTune      & 18  & 154 & 70.6\% & 2 (0.51) & 2 (0.63) & 2 (0.88) & 3 (0.58) \\
METR$^\dagger$ & 20 & 210 & 3.7\%  & = key & 3 (0.51) & 3 (0.55) & n/a$^*$ \\
MLE-bench$^\dagger$ & 27 & 74 & 10.8\% & = key & 4 (0.10) & 3 (0.52) & 3 (0.73) \\
\bottomrule
\end{tabular}
\end{table}

\paragraph{Binarize a benchmark that resolves, and it saturates.} ALE-Bench's official score, mean contest performance, separates its top ten into three tiers. Scored as an answer key against the median human, the same runs give one tier ($S=1.00$), and the objective restores three ($S=0.03$; with 40 tasks, robustly at least two). Across task resamples, the key gives one tier every time and the objective at least two 99.6\% of the time. Partial credit below the reference does not help: the capped score still gives one tier. Keeping the bit below the reference and crediting only the margin above it gives all three.

\paragraph{Where it is not saturated, the margin does not help.} On SWE-fficiency, partial credit below the reference helps most, and the official score, a harmonic mean in which failed tasks dominate, resolves worst of all (one tier): how a continuous score is aggregated matters as much as whether it is continuous. On METR, few runs beat the reference. On MLE-bench the key already separates the top ($S=0.10$), and the heavy tail of Kaggle metrics blurs it ($S=0.73$). The margin above the reference carries information mainly where the key has run out.

\section{Failure B on Terminal-Bench, and the Ignored Cost}\label{sec:c3}

\begin{figure}[tb]
\centering
\includegraphics[width=\textwidth]{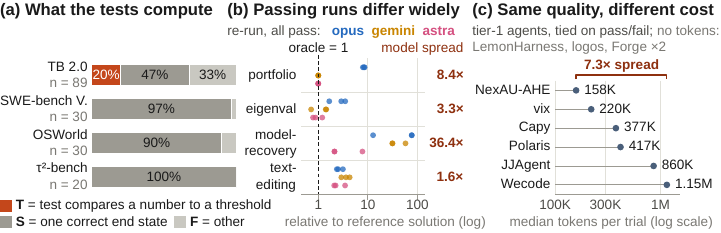}
\caption{\textbf{(a)} What each task's test checks. \textbf{(b)} Re-run: all runs of the three models that pass/fail ties, on four tasks where all pass, as the quantity relative to the reference solution (higher is better: speedup, inverse MSE or keystrokes). Right: best over worst model median. \textbf{(c)} Median tokens per run of the tied Terminal-Bench~2.0 top ten.}
\label{fig:recover}
\end{figure}

\paragraph{What the tests compute.} Consider the Terminal-Bench~2.0 task \texttt{query-optimize}: the agent must speed up a slow SQL query. The test runs the agent's query, measures its runtime, and passes it if the runtime is below a threshold. The runtime is computed and then thrown away. We coded the tests of all 89 Terminal-Bench~2.0 tasks and samples of SWE-bench Verified (30), OSWorld-Verified (30) and $\tau^2$-bench (20) as T (a quantity compared with a threshold), S (a unique correct end state) or F (other). On Terminal-Bench~2.0, 18 tasks (20\%) are T. The other samples contain no T tasks (Fig.~\ref{fig:recover}a; with samples this small, true rates up to 10--15\% cannot be excluded). For them our design offers a hardened gate and a cost meter, but no quality objective unless the task is restated around a quantity, as SWE-fficiency does~\cite{swefficiency2025}.

\paragraph{A controlled re-run.} Released logs record the quantity for only four T tasks, on differing machines, so we re-ran 13 T tasks with tests changed only to print it. The re-run is a small instance of the design of §\ref{sec:arch}: the unchanged test is the gate, the printed quantity the objective, and per-trial dollars and tokens the meter. We ran four models (gpt-5.3-codex, claude-opus-5, gpt-6-astra, gemini-3.1-pro; codex, opus, astra, gemini below), reached through a third-party OpenAI-compatible gateway, three trials per task in the Terminus-2 agent~\cite{terminalbench2026} (156 runs, about \$115 at list prices). Each run had its own CPU core. Pass/fail ties opus, astra and gemini and separates codex (within-task permutation tests, Holm).

\paragraph{Same pass, an order of magnitude apart.} The tied three differ widely on the quantity (Fig.~\ref{fig:recover}b): on \texttt{portfolio-optimization} all their runs pass, yet opus's median speedup is 10.9$\times$ and the others' 1.3$\times$, the reference's level; on \texttt{pytorch-model-recovery} the median MSE spans 0.016 to 0.60. Of all passing runs, 69\% beat the reference solution. Ranking models is harder: an anchor-free permutation test (failures at the threshold, standardized log quantity, Holm) separates four of six pairs, pass/fail three, and the one tied pair it separates (opus and gemini, Holm $p=0.031$) is not separated on passing runs alone. With three trials per cell, the robust evidence is per task. Two codex runs passed by reward hacking, more verifier error (§\ref{sec:c1}), and both stood out in the printed quantity; counted as failures, they leave codex below its leaderboard entry on these tasks (stratified $p=0.025$).

\paragraph{What the key ignores about cost.} Over all 89 tasks, the six tied top-ten agents that log tokens differ $7.3\times$ in median tokens per run (Fig.~\ref{fig:recover}c) and, for the four priced at list rates, $5.5\times$ in median cost per run (\$0.44 to \$2.47; every pair significant); the cheapest per run uses the priciest model per token. Where per-run cost is released, cost per accepted run separates pass/fail ties (4 of 8 adjacent mini-SWE-agent pairs on SWE-bench Verified and 4 of 5 priced Terminal-Bench~2.0 pairs in the top pass-rate tier differ at $p<0.05$), but under one scaffold this mostly reflects price per token and verbosity, not ability.

\section{Why Graded Credit, Not a Higher Bar, Fixes It}\label{sec:score}

On ALE-Bench a run's contest score can be compared with any bar, so we can test the usual remedy for a saturated key, a higher bar, without a new version.

\paragraph{Candidates.} Let $J$ be the quantity a task measures (runtime, contest score), $J_{\text{base}}$ a baseline, $J_{\text{ref}}$ the human or expert reference, and $s$ the run's position between them ($0$ at the baseline, $1$ at the reference). Pass/fail \emph{bars} put the target at the median human (the key), at the contest winner, or where it best separates the models of the time. Continuous scores from prior work: \emph{capped} $\min(s,1)$; \emph{linear} $(J-J_{\text{base}})/(J_{\text{ref}}-J_{\text{base}})$, the human-normalized score of reinforcement learning~\cite{mnih2015dqn}; \emph{log-ratio} $\log(J/J_{\text{base}})/\log(J_{\text{ref}}/J_{\text{base}})$, as in systems benchmarking~\cite{fleming1986lie}; the log-ratio \emph{clipped} at~$3$; and the \emph{percentile} among a pool of runs on the task~\cite{dolan2002profiles,agarwal2021precipice}. All are floored at $0$ and averaged over tasks.

\paragraph{A pass/fail bar sees one band of ability.} We split the 141 ALE-Bench models with known or estimated release dates into 46 released by the end of 2025 and 95 released later (Table~\ref{tab:score}). The median-human bar ties the later top ten. The contest-winner bar separates them into two tiers, but scores 89\% of the earlier models zero on every task (agreement with the human contest rating: $0.43$). A bar tuned to separate the earlier models puts the later top ten in one tier. Each bar resolves one band of ability; raising it moves the band instead of widening it. This is what a harder bar on the same tasks does; a real new version also adds tasks, which we do not simulate. Nor does varying the bar by task: setting each task's bar where half the earlier models pass still ties the later top ten. Several bars per task, a graded score in the spirit of polytomous item response models, do resolve them; a continuous score is the limit of that. Bounded scores with room above the reference rank both groups (clipped log-ratio, percentiles), while the linear score and the unclipped log-ratio tie the earlier top ten, which supports the clip. The clipped log-ratio agrees with humans about as well as the key (difference $+0.02$, 95\% CI $-0.002$ to $+0.05$) and is more stable across task halves ($+0.03$, CI excludes 0). On the 32 later models with verifiable dates, the tiers are unchanged (3 vs 1; $S=0.41$).

\begin{table}[tb]
\caption{ALE-Bench models released by the end of 2025 (earlier) and after (later); 63 of the later models have reconstructed dates. Tiers: among the top ten, with $S$ in brackets; with 40 tasks, counts above two depend on calibration. Human: Kendall $\tau$ with the contest rating. Zero: models scoring zero on every task. Split-half: ranking agreement across random task halves.}\label{tab:score}
\centering\footnotesize\setlength{\tabcolsep}{2.5pt}
\begin{tabular}{@{}lccccccc@{}}
\toprule
& \multicolumn{3}{c}{Earlier models (46)} & \multicolumn{3}{c}{Later models (95)} & \\
\cmidrule(lr){2-4}\cmidrule(lr){5-7}
Score & Tiers ($S$) & Human & Zero & Tiers ($S$) & Human & Zero & Split-half \\
\midrule
Bar: median human (key) & 1 (0.94) & 0.89 & 9\% & 1 (1.00) & 0.91 & 0\% & 0.81 \\
Bar: contest winner     & 1 (1.00) & 0.43 & \textbf{89\%} & 2 (0.08) & 0.61 & 73\% & 0.66 \\
Bar: tuned on earlier   & 2 (0.41) & 0.88 & 0\% & 1 (0.85) & 0.81 & 0\% & 0.78 \\
Per-task bars, calibrated & 1 (0.95) & 0.89 & 4\% & 1 (0.85) & 0.75 & 0\% & -- \\
Graded (6 bars per task) & 2 (0.94) & 0.92 & 0\% & 3 (0.15) & 0.95 & 0\% & -- \\
\midrule
Capped                  & 2 (0.75) & 0.88 & 0\% & 1 (0.93) & 0.86 & 0\% & 0.79 \\
Linear                  & 1 (0.83) & 0.87 & 0\% & 1 (0.82) & 0.87 & 0\% & 0.82 \\
Log-ratio               & 1 (0.70) & 0.90 & 0\% & 3 (0.005) & 0.92 & 0\% & 0.83 \\
\textbf{Log-ratio, clip 3} & 2 (0.68) & 0.90 & 0\% & \textbf{3 (0.03)} & \textbf{0.94} & 0\% & 0.83 \\
Percentile, earlier pool & 2 (0.87) & 0.92 & 0\% & 2 (0.60) & 0.88 & 0\% & 0.81 \\
\bottomrule
\end{tabular}
\end{table}

\paragraph{Choosing among continuous scores.} Capped scores saturate with the bar. Percentiles depend on the pool: adding or removing models flips up to a third of pairs. Linear scores are dominated by a few tasks: one ALE-Bench problem's contest score spans four orders of magnitude, and 79\% of the linear score comes from the top 5\% of tasks (log-ratio: 11\%).

\paragraph{Why the clipped log-ratio.} The log form is derived rather than chosen on ALE-Bench. A score of a positive quantity should not depend on its unit, so it is a function $f$ of $x=J/J_{\text{base}}$, and credit should add under composition, so $f(xy)=f(x)+f(y)$. The only continuous solutions of this Cauchy equation are $f(x)=k\log x$; averaging over tasks then summarizes ratio-scale results meaningfully~\cite{fleming1986lie,aczel1989merging}, and dividing by $\log(J_{\text{ref}}/J_{\text{base}})$ sets $k$ per task so that the reference scores $1$. The clip at three reference gaps is a stated convention: it bounds any single task's share of the score to $3/n$ for $n$ tasks. Without it, a task whose reference barely beats the baseline dominates: on SWE-fficiency, one with an expert speedup of 1.001 would hold a third of the score. Its value matters little: it binds on 4\% of ALE-Bench runs, and from clip 2 to none, agreement with humans stays within 0.90--0.94 in both cohorts and the tiers do not change. A clip that binds often signals a weak reference, to be replaced rather than the clip tuned. We also ablated the form: unclipped log-fold scores, raw or scaled by each task's anchor-set spread, agree less with humans (later cohort: at most 0.89 against 0.935) and are less stable; a quality--cost composite ranks cost (§\ref{sec:arch}).

We therefore propose the \textbf{log-ratio, floored at the baseline and clipped at three reference gaps, averaged over tasks}, for positive quantities such as runtime or contest score, and the linear score with the same floor and clip for quantities with no natural zero, such as Kaggle metrics. It gives at least the key's top-ten tiers on SWE-fficiency and AlgoTune, one fewer on MLE-bench.

\section{A Benchmark Design That Scores Acceptance, Quality and Cost Separately}\label{sec:arch}

\begin{figure}[tb]
\centering
\includegraphics[width=\textwidth]{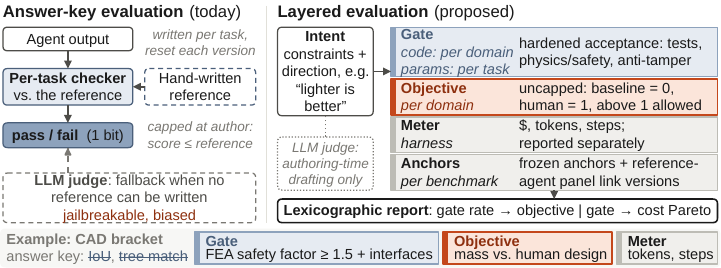}
\caption{\textbf{Left:} an answer key merges acceptance, quality and the author's solution into one per-task checker that emits one bit, with an LLM judge as the fallback. \textbf{Right:} the layered design. Quality is scored once per domain with room above the reference, and cost is metered separately.}
\label{fig:arch}
\end{figure}

We separate the roles that an answer key merges (Fig.~\ref{fig:arch}); $x$ is an agent's final artifact and $h$ its trajectory.

\paragraph{Intent} states what the human wants: hard constraints and a direction of preference (``the bracket must carry the load; lighter is better''), never a procedure or a target shape, which would force per-task keys~\cite{cadworld2026}; the bracket intent needs one physics check and one mass objective per domain.

\paragraph{Gate} $G(x)\in\{0,1\}$ is a deterministic, hardened acceptance check (behavioural, physics or safety, anti-tamper), where verifier hardening~\cite{abc2025,utboost2025,sweabs2026} belongs.

\paragraph{Objective} $s(x)$ normalizes a raw quantity $J(x)$ (runtime, mass, cost, contest score) to dated anchors, mapping $J_{\text{base}}$ to $0$ and $J_{\text{ref}}$ to $1$:
\[
s(x)=G(x)\cdot\min\!\Big(3,\ \max\!\Big(0,\ \tfrac{\log (J(x)/J_{\text{base}})}{\log (J_{\text{ref}}/J_{\text{base}})}\Big)\Big)
\]
for positive quantities, averaged over tasks (§\ref{sec:score}). The design requires a strong reference, near the best known solution, and excludes tasks whose reference barely beats the baseline by a margin fixed in advance, as SWE-fficiency does. A clip that binds often flags a weak reference, to be replaced, as on four re-run tasks. The objective is written once per domain; exposing it in the re-run took one or two lines per task.

\paragraph{Meter} $c(h)$ records dollars at a dated price table, tokens, steps and time, reported as cost-to-target~\cite{hansen2016coco} and never blended into $s$.

\paragraph{Anchors} $J_{\text{base}}$ and $J_{\text{ref}}$ are frozen with a date, so two versions of a benchmark can share one scale. Runtimes drift with hardware, so the baseline and reference artifacts are re-measured on the machine of the runs and enter only as ratios, as the re-run did. Re-running three reference agents linked an old and a harder new half of ALE-Bench within 1--3 points of the accuracy of re-running every old model, which a moving bar cannot do.

\paragraph{Judges and reporting.} LLMs may draft gate and objective code at authoring time, with human audit~\cite{opencomputer2026}, but do not score at run time, so no judge can be persuaded; this is a design position~\cite{onetoken2025}, not a finding. Results are reported lexicographically: the gate rate with statistical tiers, then the mean of $s$ (a run that fails the gate scores $0$, so the objective never rewards an invalid run). Cost is a separate axis, reported jointly with quality and its Pareto front. Any claimed lead states the interval of exchange rates $\lambda$ (folds of quality per fold of cost) over which it holds; holding for every $\lambda$ is Pareto dominance. A composite would rank cost, not quality: on ALE-Bench its $\tau$ with the human rating is $-0.35$ in the later cohort, and 129 of 137 adjacent pairs flip as $\lambda$ moves from 0.5 to 2.

\section{Limitations and Conclusion}\label{sec:concl}

\paragraph{Limitations.} Per-agent verifier error exists only for SWE-bench Verified, where we take SWE-ABS's tests as ground truth and can bound false fails only by simulation. The sliding-window result rests on ALE-Bench, with partly reconstructed dates; AlgoTune agrees weakly, and MLE-bench, whose key is not saturated, shows no such effect. The re-run's aggregate separation of tied models is borderline (three trials). Rubric-graded deliverables such as GDPval have neither a gate nor a scalar objective, and objectives can be gamed~\cite{robustkbench2025}, so the gate must stay hardened.

\paragraph{Conclusion.} Agent benchmarks often saturate in the score, not the tasks: the bit is too coarse to separate leading agents, and capped at its author. Benchmarks should build in an anchored objective with room above the reference and a separate cost meter, so a saturated key ends a version, not the measurement.

\bibliographystyle{splncs04}
\bibliography{refs}
\end{document}